\documentclass[runningheads]{llncs}
\usepackage[T1]{fontenc}
\usepackage{graphicx}
\usepackage{booktabs}
\usepackage[table,xcdraw]{xcolor}
\usepackage{caption} 
\newcommand*\rot{\rotatebox{80}}
\usepackage[title]{appendix}

\begin{document}
\title{A No-Code Low-Code Paradigm for Authoring Business Automations Using Natural Language} 

\titlerunning{Authoring Business Automations Using Natural Language}
%
%
\author{Michael Desmond \and
Evelyn Duesterwald \and
Vatche Isahagian \and \\
Vinod Muthusamy}
%
\authorrunning{Desmond, Duesterwald, Isahagian, Muthusamy}
%

\institute{IBM Research AI, USA \\
\email{\{mdesmond,duester\}@us.ibm.com, vatchei@ibm.com, vmuthus@us.ibm.com}
}
\maketitle              

\begin{abstract}

Most business process automation is still developed using traditional automation technologies such as workflow engines. These systems provide domain specific languages that require both business knowledge and programming skills to effectively use. As such, business users often lack adequate programming skills to fully leverage these code oriented environments. We propose a paradigm for the construction of business automations using natural language. The approach applies a large language model to translate business rules and automations described in natural language, into a  domain specific language interpretable by a business rule engine. We compare the performance of various language model configurations, across various target domains, and explore the use of constrained decoding to ensure syntactically correct generation of output.

\end{abstract}

\section{Introduction}
Business process automation (BPM) has emerged as a multi-billion dollar industry \cite{gartnerforecastsautomation} driven by the adoption of digital transformation and IT automation in the enterprise. Trends in automation are further accelerating due to changes in the nature of work as a result of the pandemic. Much of this automation has been developed using traditional automation technologies, such as workflow engines and operational decision management services. Usually, these technologies come with a custom automation language, such as BPMN~\cite{aagesen2015bpmn} and DMN~\cite{biard2015separation}, that require users to have adequate programming and infrastructure skills to develop, debug and deploy. As enterprises continue to embrace automation, there is a growing need to enable business users (citizen developers) to build automations of their mundane business tasks. Citizen developers are experts in their fields, but often lack adequate programming skills to code and debug automations using languages and tools designed for proficient developers.

Multiple low-code efforts to ease the development burden for citizen developers have been proposed. Power Automate~\cite{powerautomate2020} provides a visual tool that enables users to create automations. Robotic Process Automation (RPA) vendors such as UIPath, Automation Anywhere, and IBM RPA provide recorders that enable users to record their actions and automate mundane tasks. While it is easier to build automations using these low code tools, they come with their own set of challenges~\cite{leno2021robotic}. Furthermore, the generated RPAs are typically limited in execution scope, and still require advanced development skills to modify and maintain. 

Recent advances in AI in domains such as AI planning \cite{chakraborti2020d3ba}, and Natural language Processing (NLP) \cite{rizk2020conversational} have enabled the construction of more advanced automations. While such approaches have shown great promise for building sophisticated flows that go beyond simple mundane task automation, there has been less focus on
using AI to lower the skill burden for citizen developers. 

We believe that lowering the barrier of entry for business automation may provide an even larger opportunity for AI.   Specifically, we advocate the use of natural language (NL) as an interface for ease of development and maintenance of business automations. If citizen developers can express their automation requirements using familiar NL, there is no skill hurdle to overcome. However, constructing NL interfaces also poses unique challenges. NL is inherently ambiguous making the mapping of user provided NL specifications to executable automation code very challenging, 

In this vision paper, we propose a paradigm for the construction of effective no-code interfaces for authoring business automations using NL. Our approach leverages recent advances in NLP to offer citizen developers an NL-based authoring experience while still providing effective translation into non-ambiguous executable automation code.





\section{Proposed Paradigm}

\begin{figure}[tb]
    \centering
    \includegraphics[width=0.98\linewidth]{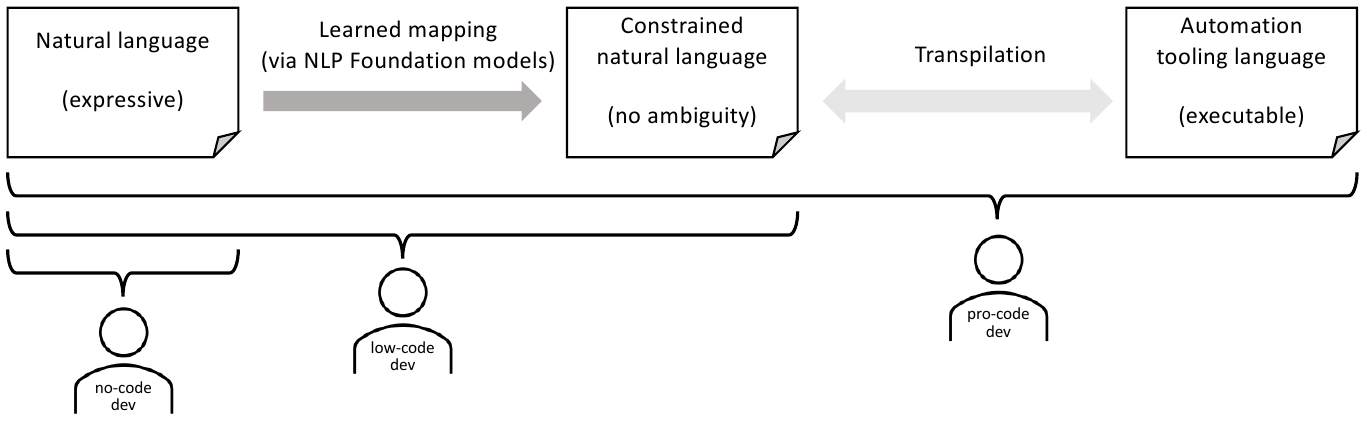}
`    \caption{NL-based no-code/low-code solution for building automations.}
    \label{fig:approach}
\end{figure}

\begin{figure}[tb]
    \centering
    \includegraphics[width=0.98\linewidth]{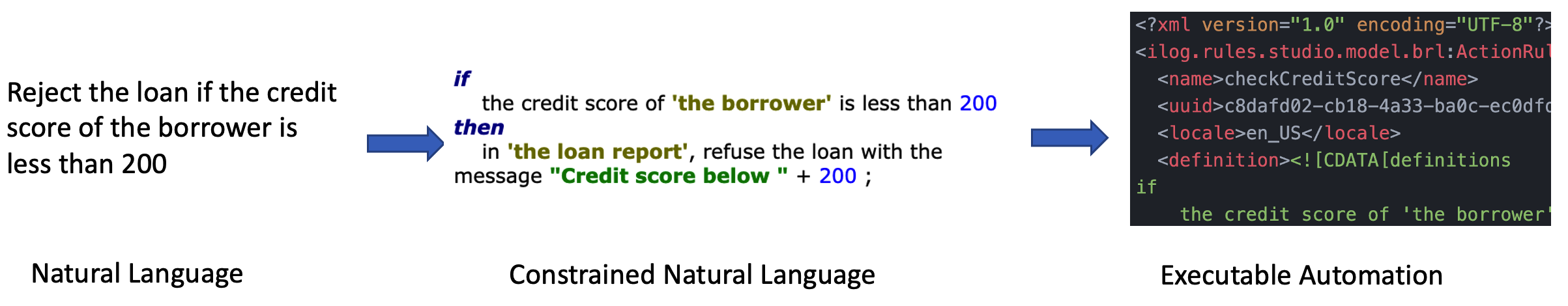}
    \caption{NL, CNL and automation code example in Operational Decision Manager (ODM)~\cite{ibmodm}. The NL expression is hypothetical and not supported in  ODM.}
    \label{fig:example}
\end{figure}

Our paradigm for NL based automation is illustrated in Figure~\ref{fig:approach}. As shown in the figure, we connect an NL based no-code front end to a target automation language backend via the introduction of an intermediate Constrained Natural Language (CNL). The CNL bridges the semantic gap between the citizen developer's NL input specification and the low-level automation code.  CNLs are non-ambiguous domain specific programming languages (DSLs) but CNL domain keywords and abstractions bear more resemblance to NL. Due to their resemblance to NL, code written in a CNL tends to be easy to read and understand. However, learning to write code in a CNL is conceptually not any easier than learning other DSLs. Thus, CNL code is generally not easy to write from scratch. Figure~\ref{fig:example} provides an example illustrating the NL, CNL, and automation code representation of the same business rule example. 
 
The intermediate CNL provides several important advantages:
First, it enables the rapid construction of learned  NL to CNL mappings based on emerging technology in language foundation models.
Second, as an easy-to-read language, it provides a user-friendly interface for the citizen developer to review and validate whether they achieved what they had in mind.
Third, as a DSL, it facilitates the construction of  source-to-source transformation (transpilation) to the underlying automation code using traditional compiler technology~\cite{kulkarni2015transpiler}. 


Our approach provides a combined no-code/low-code authoring experience where citizen developers use NL to build automations that can be reviewed and validated at the CNL level before being transpiled to the underlying automation code.  Another advantage targets proficient developers: they can use the NL front end to quickly bootstrap an initial automation that can further be edited and refined at the automation code level. 

In order to construct the NL front end, we need to the ability to quickly learn a translation from the NL input to the corresponding representation in the CNL. We report on early experience with constructing these learned mappings using emerging NLP foundation models with minimal training data~\cite{foundationmodels}. Our preliminary results, summarized in Section~\ref{sec:experiments} demonstrate both the feasibility and practicality of the proposed paradigm for building NL based interfaces for business automation.

%


\section{Preliminary Experiments} \label{sec:experiments}

We evaluated the efficacy of NL to CNL translation using generative language models, comparing a variety of language model (LM) configurations and the use of constrained decoding~\cite{shin2021constrained}, fine-tuning vs. prompting~\cite{liu2021pre}, and varying availability of training data. We experimented with two distinct data sets: a business rules dataset, based on the miniloan ruleset\footnote{\url{https://github.com/DecisionsDev/odm-for-dev-getting-started}}, which included a set of CNL business rules with corresponding NL paraphrases; and the publicly available overnight data set~\cite{wang2015building} containing cross-domain CNL style queries with NL paraphrases provided by MTurk workers.

The task was to map NL statements into a corresponding CNL form. The mapping was learned either by fine-tuning the model, or by prompting the model at inference time. We compared the following LM configurations:
T5
~\cite{raffel2019exploring}, BART
~\cite{lewis2019bart}, as well as 
gpt-neo-125M, 
gpt-neo-1.3B, and 
gpt-neo-2.7B\footnote{https://huggingface.co/EleutherAI/gpt-neo-2.7B} each corresponding to the GPT~\cite{brown2020language} architecture but with a larger parameter set.
While T5 and BART were fine-tuned using training data, the gpt models were prompted at inference time with NL-CNL samples from the same training data. We implemented a prompt generation pipeline to generate a prompt based on the similarity of an input NL statement to existing NL statements in the training data. Similar NL-CNL pairs are appended to the prompt until the context window supported by the LM architecture is full, leaving sufficient room for the desired output. 

In some LM configurations we adopt constrained decoding techniques~\cite{shin2021constrained} to restrict the generation process of the language models to adhere to a predefined grammar or language structure. We implemented constrained decoding using a prefix tree, constructed from tokenized examples of valid CNL statements. At each generation step the LM queries the prefix tree to select a valid next output token, thereby ensuring a syntactically correct CNL output.

Due to the effort involved in acquiring NL-CNL training examples we also evaluated with a reduced training set of only 100 samples. The primary metric we report on is accuracy, which counts the number of exact matches generated by the LM compared to reference CNL statements. Given the semantics of logical expressions (and, or) which are contained in both data sets, exact matching may be underestimating performance as operands can switch order without changing semantics.   However we believe accuracy, as a conservative measure, is still a practical metric.
Additional metrics are reported in the appendix.

\smallbreak
{\noindent \bf Experimental Protocol:} Each of the data sets 
were split into 70\% training, 24\% test and 6\% validation sets. 
A smaller training set containing 100 examples was also sampled from the initial training set to simulate a limited training environment. In the case of T5 and BART the training set was used to fine-tune instances of these models for each of the data sets. Another version of T5 and BART was also fine-tuned using the smaller 100 sample training set. 
The LM configurations were then evaluated on the NL-to-CNL translation task using common test sets. Note that the prompted models were limited to generating prompts from the same training data used to train the fine-tuned models. During constrained decoding the prefix tree was constructed using all of the available CNL examples, from the train, test and validation sets combined. The predictions from each of the pipelines were then evaluated against the reference CNL statements from the test set to produce results. All experiments were executed with one v100 GPU. Decoding was performed using four beams.

\smallbreak
{\noindent \bf Experimental Results:} Our results (summarized in Table~\ref{tab:results-acc}) indicate that NL-to-CNL translation using large language models is a promising approach. On the business rules dataset (miniloan), fine-tuned ML configurations performed better than prompted gpt configurations. 
This observation held even when comparing fine-tuned LM configurations with limited training data to prompted LMs having access to the entire training set.

Overall T5 with constrained decoding performed best, followed closely by BART with constrained decoding. Constrained decoding provided slight performance improvements over fine-tuned LM configurations with full training data, but as the performance of the base LM deteriorated (with limited training data or prompted LM configurations), the support provided by constrained decoding became more noticable. This indicates that as the translation task becomes more difficult, the value of constrained decoding becomes more apparent. 

Using limited training data (100 examples) diminished performance across all LM configurations, but the fine-tuned LMs still performed well in this setting. This is promising as it indicates that reliable translation can be achieved with a small amount of training data. Inference time became progressively slower with the larger gpt LMs, taking ~5 seconds per example with the largest parameter configuration.

\begin{table}[htp]
\small
\centering
\begin{tabular}{|l|l|l|l||l|l|l|l|l|l|l|l|l|l|l|}
\hline
\multicolumn{1}{|r|}{\textbf{Dataset family}} & \multicolumn{3}{c||}{\textbf{miniloan}} & \multicolumn{11}{c|}{\textbf{overnight}} \\
\hline
\multicolumn{1}{|r|}{\textbf{Dataset}}
& \multicolumn{2}{|c|}{\rot{\textbf{miniloan-full}}}
& \rot{\textbf{miniloan-100}}
& \rot{\textbf{basketball}}
& \rot{\textbf{blocks}}
& \rot{\textbf{calendar}}
& \rot{\textbf{calendarplus}}
& \rot{\textbf{geo880}}
& \rot{\textbf{housing}}
& \rot{\textbf{publications}}
& \rot{\textbf{recipes}}
& \rot{\textbf{regex}}
& \rot{\textbf{restaurants}}
& \rot{\textbf{socialnetwork}} \\ 
\hline
\multicolumn{1}{|r|}{\textbf{Metric}} & \textbf{INF} & \multicolumn{13}{c|}{\textbf{ACCURACY}} \\ \hline \hline

\textbf{T5} & .23 & .98 & .91 & .86 & .50 & .61 & .45 & .74 & .51 & .63 & .72 & .43 & .56 & .70 \\ \hline
\cellcolor[HTML]{FFFFC7}\textbf{T5/C.} & \cellcolor[HTML]{FFFFC7}\textbf{.29} & \cellcolor[HTML]{FFFFC7}\textbf{.99} & \cellcolor[HTML]{FFFFC7}\textbf{.98} & .87 & .50 & .67 & .63 & .75 & .54 & .64 & .72 & .43 & .58 & .71 \\ \hline
\textbf{BART} & .12 & .98 & .91 & {\color[HTML]{000000} .85} & {\color[HTML]{000000} .47} & {\color[HTML]{000000} .65} & .43 & .72 & .54 & .59 & .70 & .73 & .57 & .71 \\ \hline
\textbf{BART/C.} & .16 & .98 & .97 & {\color[HTML]{000000} .85} & {\color[HTML]{000000} .48} & {\color[HTML]{000000} .65} & .57 & .72 & .55 & .59 & .70 & .74 & .57 & .72 \\ \hline \hline
\textbf{GPT125M} & .47 & .16 & .12 & .34 & .07 & .15 & .04 & .30 & .09 & .23 & .23 & .09 & .18 & .22 \\ \hline
\textbf{GPT125M/C.} & .50 & .49 & .35 & .35 & .11 & .20 & .09 & .38 & .14 & .28 & .28 & .15 & .21 & .25 \\ \hline
\textbf{GPT1.3B} & 2.6 & .50 & .25 & .53 & .24 & .33 & .13 & .54 & .29 & .36 & .36 & .30 & .37 & .34 \\ \hline
\textbf{GPT1.3B/C.} & 2.6 & .75 & .54 & .54 & .26 & .34 & .16 & .57 & .32 & .36 & .37 & .33 & .43  & .35 \\ \hline
\textbf{GPT2.7B} & 4.9 & .43 & .26 & .60 & .28 & .40 & .10 & .54 & .33 & .40 & .43 & .38 & .41 & .38 \\ \hline
\textbf{GPT2.7B/C.} & 5.0 & .76 & .63 & .63 & 34 & .46 & .21 & .60 & .37 & .42 & .45 & .44 & .45 & .41 \\ \hline
\end{tabular}

\caption{Accuracy (exact match prediction) across language models and data sets. 
Each LM was run without and with constrained decoding (indicated with \textbf{/C.}). Miniloan-100 indicates the limited 100 examples training set was used. INF indicates inference time in seconds (only shown for miniloan-full). The best LM configuration is highlighted.}

\label{tab:results-acc}
\end{table}





\section{Discussion}


The experiments above offer encouraging results for quickly building an NL no-code interface for citizen developers to author automations. We discuss a few open challenges to make this approach more practical.

One challenge is to design the CNL. Some automation tools~\cite{ibmodm} already have a pre-defined CNL but others will require a developer to define a grammar and ontology for the CNL that ensures that the CNL is easy to read, and enables an unambiguous mapping to the target automation language. Compiler technologies can then be used to generate the transpilation code~\cite{kulkarni2015transpiler}. 

Another related challenge is the creation of training examples needed for the NL-to-CNL translation. Our use of large pre-trained language models~\cite{foundationmodels} reduces but does not entirely eliminate the need for a training corpus. It is still an open question how large the training set needs to be. In our experiments we achieve strong results with 100 training examples. We expected that further advances in few-shot learning may help to further reduce the need for training~\cite{perez2021true}.

Constrained decoding using a prefix tree is well-suited to generate output according to a predefined corpus of CNL examples, but does not account for user defined variables and literals. For example, in the CNL expression \texttt{customer age is greater than 18 and customer credit score is more than 600}, 
the values \texttt{18} and \texttt{600} 
may not appear in the CNL corpus and thus pose a problem during constrained decoding. A context-aware algorithm might solve this problem by extracting variables and literals from CNL inputs and replacing such values with special marker tokens during a pre-processing step. When such a marker token is predicted during decoding, the LM can be presented with the selection of extracted context values in place of the marker. We leave this problem of generalizing constrained decoding to future work. 

We are currently also exploring a number of ways to generalize and extend our approach. 
For example the use of a CNL as an intermediate language in Figure~\ref{fig:approach} is primarily motivated by our use of pre-trained language models, which work best when the target language is close to NL. In general, the intermediate language can be any DSL, such as a YAML representation of a visual diagram, or DSLs based on Python.  Such DSLs would require different pre-trained models than the ones we have experimented with. Conceptually, the only requirement is that the language is easy to understand, review and validate by the target user. 
Another extension to our approach is to consider multi-modal interfaces. For example, citizen developers may draw a sketch of their desired automation, verbally describe a decision rule, markup a procedure document with edits, demonstrate the steps of a workflow, or any combination of these modalities. The most productive combination of input modalities may depend on both the end user as well as the business automation domain. 

\section{Conclusions}

In this paper we presented a paradigm for the construction of effective and efficient NL interfaces to authoring business automations.  Our paradigm provides a combined  no-code/low-code environment where users author automations in natural language which then will be translated into a consumable CNL form that can be reviewed and validated before being transpiled into automation code of the underlying business platform. Early experimental results demonstrate that our no-code/low-code paradigm provides a viable and practical approach to constructing NL interfaces for automation, requiring only 100 training samples and off-the-shelf pre-trained language models. As such, our approach makes an important contribution towards a vision of enhanced productivity for business users through the use of interfaces that most naturally model the way they reason about their business domain and automation.



%
%
\bibliographystyle{splncs04}
\bibliography{bibtex}

\appendix

\appendix
\label{appendix}
\section{Additional Results}
\begin{table}[]
\centering
\begin{tabular}{@{}|l|l|l|l|l|l|l|l|@{}}
\toprule
\textbf{LM}    & \textbf{BLEU} & \textbf{RGL} & \textbf{ACC} & \textbf{BLEU100} & \textbf{RGL100} & \textbf{ACC100} & \textbf{INF} \\ \midrule
T5             & .99           & .99          & .98          & .98              & .98             & .91             & .23          \\ \midrule
\rowcolor[HTML]{FFFFC7} 
\textbf{T5/C.} & \textbf{.98}  & \textbf{.99} & \textbf{.99} & \textbf{.98}     & \textbf{.99}    & \textbf{.98}    & \textbf{.29} \\ \midrule
BART           & .99           & .99          & .98          & .98              & .98             & .91             & .12          \\ \midrule
BART/C.        & .99           & .99          & .98          & .98              & .98             & .97             & .16          \\ \midrule
GPT125M        & .41           & .64          & .16          & .33              & .62             & .12             & .47          \\ \midrule
GPT125M/C.     & .63           & .77          & .49          & .48              & .69             & .35             & .50          \\ \midrule
GPT1.3B        & .75           & .85          & .50          & .58              & .77             & .25             & 2.6          \\ \midrule
GPT1.3B/C.     & .90           & .92          & .75          & .73              & .83             & .54             & 2.6          \\ \midrule
GPT2.7B        & .74           & .84          & .43          & 0.63             & .81             & .26             & 4.9          \\ \midrule
GPT2.7B/C.     & .91           & .93          & .76          & .83              & .89             & .63             & 5.0          \\ \bottomrule
\end{tabular}
\caption{Full experimental result on a business rules (miniloan) dataset. The first column indicates the LM configuration. Each LM was run with and without constrained decoding, indicated with \textbf{/C.} following the LM name. 
BLEU*, RGL* and ACC* columns indicate BLEU, ROUGEL and accuracy for the corresponding LM config. Columns with the 100 suffix indicate the same scores but with a limited training set of 100 examples. INF indicates the inference time (seconds). }
\label{tab:results-full-miniloan}
\end{table}

\end{document}